\documentclass[runningheads]{llncs}

\usepackage{eccv}

\usepackage{eccvabbrv}
\usepackage{wrapfig} 
\usepackage{graphicx}
\usepackage{booktabs}
\usepackage{float}
\usepackage{placeins}
\usepackage[accsupp]{axessibility}  % Improves PDF readability for those with disabilities.

\usepackage{xcolor} % 用于颜色
\usepackage{pifont} % 用于对错符号
\usepackage{multirow}
\usepackage{float}
\usepackage{hyperref}

\usepackage{pifont}  
\usepackage[table,dvipsnames]{xcolor}
\usepackage{subcaption} 
\definecolor{MyGreen}{RGB}{0,128,0}

\usepackage{orcidlink}
\newcommand{\equalcontrib}{\ensuremath{^\dagger}}
\newcommand{\correspondingauthor}{\ensuremath{^*}}

\begin{document}

\title{Driving Like Yourself: A Benchmark for Closed-Loop Personalized End-to-End Autonomous Driving} 

\titlerunning{Driving Like Yourself}

% Second Author\inst{2,3}\orcidlink{1111-2222-3333-4444} \and
% Third Author\inst{3}\orcidlink{2222--3333-4444-5555}}

\author{
Xiaoru Dong\inst{1,2}\orcidlink{0009-0000-6009-9055}\equalcontrib \and
Ruiqin Li\inst{2}\orcidlink{0009-0008-6853-5382}\equalcontrib \and
Xiao Han\inst{2}\orcidlink{0009-0000-8543-6345}\equalcontrib \and
Zhenxuan Wu\inst{2}\orcidlink{0009-0001-1412-7331} \and
Jiamin Wang\inst{2}\orcidlink{0009-0006-7085-0301} \and
Jian Chen\inst{1}\orcidlink{0000-0002-4570-2271} \and
Qi Jiang\inst{2}\orcidlink{0009-0007-4310-4781} \and
Siu Ming Yiu\inst{1}\orcidlink{0000-0002-3975-8500}\correspondingauthor \and
Xinge Zhu\inst{3}\orcidlink{0000-0003-0107-8099} \and
Yuexin Ma\inst{2}\orcidlink{0000-0001-7237-988X}\correspondingauthor
}

\authorrunning{Dong et al.}

\institute{
The University of Hong Kong\\
\and
ShanghaiTech University\\
\and
The Chinese University of Hong Kong\\
\email{xrdong@cs.hku.hk, mayuexin@shanghaitech.edu.cn}
}

% Springer Heidelberg, Tiergartenstr.~17, 69121 Heidelberg, Germany

\maketitle

\begingroup
\renewcommand\thefootnote{}
\footnotetext{\correspondingauthor\ Corresponding authors.}
\footnotetext{\equalcontrib\ Equal contribution.}
\endgroup

\begin{abstract}
% \begin{abstract}

\vspace{-1ex}
Human driving behavior is inherently diverse, yet most end-to-end autonomous driving (E2E-AD) systems learn a single average driving style, neglecting individual differences. Achieving personalized E2E-AD faces challenges across three levels: limited real-world datasets with individual-level annotations, a lack of quantitative metrics for evaluating personal driving styles, and the absence of algorithms that can learn stylized representations from users’ trajectories. To address these gaps, we propose Person2Drive, a comprehensive personalized E2E-AD platform and benchmark. It includes an open-source, flexible data collection system that simulates realistic scenarios to generate scalable, diverse personalized driving datasets; style vector–based evaluation metrics with Maximum Mean Discrepancy and KL divergence to comprehensively quantify individual driving behaviors; and a personalized E2E-AD framework with a style reward model that efficiently adapts E2E models for safe and individualized driving. Crucially, our framework enables plug-and-play personalization by fine-tuning only the trajectory prediction head, preserving the pretrained base model and ensuring safety. Extensive experiments demonstrate that Person2Drive enables fine-grained analysis and effective personalization, while preserving driving performance and success rate even in challenging scenarios. 

\vspace{-0.5ex}
% \end{abstract}
\end{abstract}

\section{Introduction}
\label{sec:intro}

Traditional autonomous driving systems rely on modular pipelines that separate perception, prediction, and control, often leading to suboptimal coordination, whereas recent end-to-end (E2E-AD) approaches use unified deep learning models to map sensor inputs directly to control outputs, achieving notable improvements in behavioral consistency and safety \cite{gruyer2017perception,chib2023recent,chen2024end}. However, most E2E-AD approaches employ a single model to fit all drivers, effectively learning an average driving style while neglecting individual differences. In reality, human driving behavior is inherently diverse—shaped by cognitive patterns, habits, and personality traits \cite{liao2024review}. Overlooking such individuality limits our understanding of human decision-making and impedes the development of human-centered, trustworthy driving systems \cite{holzinger2022digital}. Personalized E2E-AD, by contrast, aims to adapt driving behavior to individual preferences and traits, thereby improving comfort, trust, and user acceptance \cite{hasenjager2017personalization,hauslschmid2017supportingtrust,hang2020human}.

To address personalization in autonomous driving, existing methods can be roughly divided into three categories. Early approaches introduce rule-based adjustments or trajectory-level tuning on modular pipelines \cite{hasenjager2019survey}, which are inherently incompatible with end-to-end learning. With the rise of E2E-AD, some studies condition models on discrete driving-style labels (e.g., aggressive, normal, conservative) \cite{hao2025styledrive}. However, human driving behaviors vary continuously across scenarios, making such coarse categorization insufficient for capturing individual dynamics. More recently, language-conditioned control has been explored for interpretable human–vehicle interaction \cite{cui2024personalized,xu2025towards,qin2025contextual}, yet language signals alone cannot represent fine-grained, continuous control characteristics. This calls for a new personalized E2E-AD paradigm capable of adapting driving styles directly from users' trajectory data.
\vspace{-0.7mm}
\begin{figure*}[!t]
    \centering
    \includegraphics[width=1\linewidth]{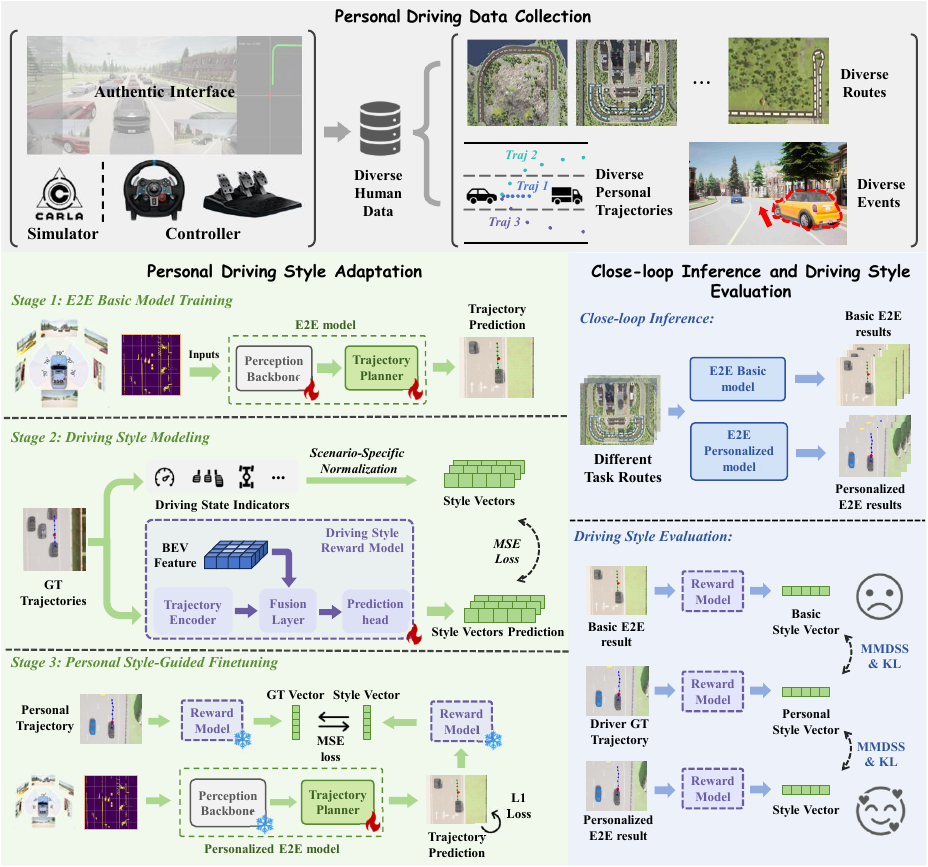}
    % \vspace{-6mm}
    \caption{\textbf{Person2Drive framework.} We develop a low-cost, scalable platform for data collection, featuring a navigation map and multi-camera views that mimic real-world driving habits, enabling authentic data acquisition across diverse scenarios, road segments, and driver styles. To achieve personalized driving style adaptation, we propose a three-stage adaptation method. Finally, the adapted end-to-end model is deployed in a simulated environment for inference, and the style model is used to evaluate the effectiveness of personalization.}
    \label{fig:framework}
    \vspace{-6ex}
\end{figure*}

Against this background, significant challenges remain in achieving truly individual-level personalized E2E-AD. These challenges arise throughout the entire process of this task, including data, evaluation, and algorithms. The challenges can be summarized as follows:
\vspace{-1.5ex}
\begin{enumerate}
    \item[$\bullet$] \textbf{Data level:} Existing driving datasets face several limitations. Many are collected in simulated or non-human environments, while real-world datasets lack individual-level style annotations. Furthermore, they rarely support data augmentation or closed-loop evaluation, limiting their applicability to personalized E2E-AD research.
    \item[$\bullet$] \textbf{Evaluation level:} Current research lacks standardized and quantitative metrics for assessing individual driving styles, leaving a gap in measuring and comparing behavioral differences across drivers.
    \item[$\bullet$] \textbf{Algorithm level:} Existing approaches struggle to learn stylized representations directly from users’ trajectory data and lack a unified end-to-end framework for personalized autonomous driving.
\end{enumerate}
\vspace{-1.5ex}

Therefore, to address these three levels of challenges, we built a comprehensive personalized E2E-AD platform and benchmark, which is called \textbf{Person2Drive}. Our framework establishes a closed-loop research pipeline that enables data collection, model training, and quantitative evaluation of individualized driving behaviors. This platform not only provides the first unified foundation for studying personalized E2E-AD but also facilitates systematic comparison and reproducible research in this emerging area. Specifically, to tackle the aforementioned challenges, our work makes the following key contributions:
\vspace{-1.5ex}

\begin{enumerate}
    \vspace{-0.5ex}
    \item[$\bullet$] \textbf{Personal driving data collection.}  We develop an open-source and highly extensible data collection platform that simulates realistic driving scenarios at low cost, enabling the construction of large-scale and easily expandable personalized driving datasets. The dataset captures individual differences in driving styles across diverse scenes and events, supporting fine-grained analysis at the individual level. Moreover, the platform allows closed-loop validation, which is crucial for evaluating E2E-AD.
    
    % We develop an open-source and extensible data collection platform that simulates realistic driving scenarios at low cost to construct a large-scale, diverse personalized driving dataset. The dataset captures individual differences in driving styles across various scenes and events, enabling fine-grained analysis at the individual level. Moreover, the platform supports closed-loop validation, which is crucial for evaluating E2E-AD.
    % \vspace{-0.5ex}
    \item[$\bullet$] \textbf{Driving style evaluation.} We comprehensively consider the dimensions in which individual driving styles manifest, including vehicle control parameters and driver behavior data. Based on these, we design style vector indicators that capture the most significant inter-individual differences and explore the use of Maximum Mean Discrepancy (MMD) and KL divergence to quantify driving style discrepancies. We further propose an MMD similarity score (MMDSS), providing the first quantitative analysis of personalized E2E-AD.
    
    \item[$\bullet$] \textbf{Personal driving style adaptation.} 
    %Considering preserving the performance of the base E2E model while enabling personalization, we introduce a personalized training framework and a style reward model based on style vector indicators to learn stylized representations from users’ driving trajectories. Using this reward model, we fine-tune only the final trajectory prediction head, achieving efficient and safe personalized E2E-AD, providing a practical and general solution for individualized E2E-AD.
    Aiming to maintain the performance of the baseline E2E model while realizing personalization, we introduce a dedicated personalized training framework and a style reward model built upon style vector indicators, which extracts stylized representations from users’ driving trajectories. By employing this reward model, we exclusively fine-tune the final trajectory prediction head, enabling efficient and robust personalized E2E-AD. 
    
\end{enumerate}
\vspace{-1ex}

\vspace{-1ex}
\section{Related Work}
\vspace{-1ex}
% However, they still do not take into account the specific individual-level preferences and styles of users. 
% Although StyleDrive designs a new framework for learning style-conditioned E2EAD models, it is still limited to classifying different users into similar styles, and it is impossible to truly capture the specific style at the individual level of each user. To address these limitations, we propose Person2Drive, which models each user in a unique style, and uses a pre-training and fine-tuning framework to obtain an E2EAD model for effective vehicle-end deployment.

\subsection{End-to-End Autonomous Driving}
\vspace{-0.5ex}
E2E-AD directly maps raw sensor inputs to vehicle control commands using unified models, simplifying system design and enabling joint optimization across perception, prediction, and control for more consistent and robust driving. This paradigm not only advances end-to-end autonomous driving systems but also provides a new perspective for modeling and understanding complex driving behaviors. Codevilla et al. \cite{codevilla2018end} demonstrated the practical feasibility of this approach. Recently, transformer-based methods have shown strong capabilities in temporal reasoning and multi-modal fusion, highlighting their potential for E2E-AD \cite{prakash2021multi,jia2023think,shao2023safety}. In parallel, the development of generative and world models has enhanced the generalization and interpretability of E2E-AD \cite{li2024enhancing,zheng2024genad,jia2025drivetransformer}. Notably, DiffusionDrive introduces diffusion models into E2E-AD, achieving state-of-the-art driving performance with relatively low computational cost \cite{liao2025diffusiondrive}. Collectively, these advances not only consolidate the practical foundation of E2E-AD but also provide the technical basis for personalized driving and user-specific adaptation.

\vspace{-2ex}

\subsection{Personalized Autonomous Driving}
\vspace{-1ex}

Personalized autonomous driving (PAD) aims to adapt vehicle behavior to individual driver preferences \cite{eboli2017drivers}. Although E2E-AD has advanced significantly, most existing PAD methods are not designed for fully end-to-end deployment. Early studies focused on specific driving behaviors, such as car-following \cite{zhao2022personalized}, lane-changing \cite{liao2023driver}, and ramp-merging \cite{li2023personalized}, using techniques like Inverse Reinforcement Learning (IRL), digital twins, or context encoding. More recently, large language models (LLMs) have been incorporated to enable interpretable control, for example, D2E translates user intentions into executable instructions \cite{cui2024personalized}, and PADriver combines streaming frames with personalized textual prompts \cite{kou2025padriver}. MAVERIC proposes a data-driven approach for personalized driving \cite{schrum2024maveric}, while StyleDrive leverages a fine-tuned vision-language model to realize end-to-end personalized driving, fusing scene topology with dynamic agent interactions \cite{hao2025styledrive}. Despite these advances, existing methods remain limited: style labels are coarse-grained, trajectory-level fine-grained style representations are not learned, and a general end-to-end framework for safe and efficient personalization is still lacking.

\vspace{-2ex}

\subsection{Personalized Driving Datasets}
Personalized autonomous driving has attracted considerable attention, leading to the development of several personalized driving datasets. Early datasets explored individual differences but primarily focused on open-loop evaluation, limiting their applicability to end-to-end systems \cite{zhao2022personalized,liao2023driver,li2023personalized,ke2024d2e,wei2025pdb}. Bench2Drive provides closed-loop evaluation, yet its data is not collected from human drivers \cite{jia2024bench2drive}. StyleDrive is the first end-to-end driving dataset with style annotations, but it is semi-closed-loop and restricted to only three style categories: aggressive, normal, and conservative \cite{hao2025styledrive}. These limitations highlight the urgent need for a personalized driving dataset that supports end-to-end closed-loop evaluation, enabling true individual-level research in personalized E2E-AD.

% Some existing works have already achieved style annotation for driving datasets, which can be divided into Human-in-the-Loop settings and real-world scenarios.  Existing human-in-the-loop settings focus mainly on open-loop, which can not effectively reflect the interaction with the environment \cite{zhao2022personalized,liao2023driver,li2023personalized,ke2024d2e}. UAH-DriveSet concentrates on 3 different behaviors (normal, drowsy, and aggressive) but not real driver styles \cite{romera2016need}. StyleDrives is the first end-to-end driving dataset with a style label annotation, but it is semi-closed-loop, and the styles are limited to three categories: aggressive, normal, and conservative \cite{hao2025styledrive}. PDB isolates driver behavior by maintaining consistent external conditions, making changes in driving patterns mainly originate from the driver themselves, but not end-to-end and closed-loop \cite{wei2025pdb}. Furthermore, PDB dataset also overlook the fine-grained intra-individual difference in different scenarios. Therefore, there is an urgent need for a scene-aware personal-level closed-loop dataset and benchmark for intelligent driving style, so as to provide an evaluation benchmark and data support for the research and development of personalized end-to-end autonomous driving systems. 

\vspace{-1.5ex}

\section{Personalized Driving Benchmark}
\vspace{-1.5ex}

We introduce \textbf{Person2Drive}, a benchmark for exploring personalized E2E-AD, which emphasizes the fidelity and consistency of \textbf{human driving styles}. It comprises an open-source system for collecting multi-modal personalized driving data, a fully annotated dataset with driver identity labels, interpretable driving style evaluation, and a personalized driving style adaptation method for efficient imitation and transfer of individual driving behaviors. 
% Building upon these, we establish a comprehensive benchmark for personalized E2E-AD, as illustrated in 
The overall framework is illustrated in Figure~\ref{fig:framework}.

\vspace{-1.5ex}

\subsection{Personal Driving Data Collection}
We develop a low-cost, scalable, and open-source data collection system that supports \textbf{closed-loop end-to-end operation}, captures authentic human driving styles, and enables diverse data acquisition. With this system, we further construct a personalized driving dataset with multi-modal data.

% Collecting large-scale personalized driving data is challenging due to the need for real human participation, synchronized multi-modal sensing, and controlled yet diverse environments that ensure comparability across drivers.   
% Existing datasets mostly consist of expert-agent-generated or pre-recorded data without \textbf{identity labels,} limiting personalized driving research. 
% To overcome these limitations, we develop a scalable and reproducible \textbf{data collection system} that allows real human drivers to operate in diverse simulated environments with synchronized multi-modal recording.   
% Building upon this system, we construct \textbf{Person2Drive}—a fine-grained personalized driving dataset designed to capture human-specific driving styles.

\noindent\textbf{Data Collection System}
%closed-loop manner后要加引用
Existing closed-loop E2E-AD datasets are mostly from expert drivers and lack standardized, extensible collection frameworks, limiting research on individual driving styles. Real-world personalized data is costly and hard to control, confounding style variations with environmental factors. Therefore, there is an urgent need for a low-cost, scalable data collection system. We developed an open-source data collection system based on CARLA \cite{carla2023leaderboard}, enabling human drivers to operate under diverse traffic, scene, and weather conditions while automatically recording multi-modal personalized driving data.  
Our framework extends the implementation provided by Bench2Drive~\cite{jia2024bench2drive}, and can be easily adapted to collect additional data. 
1) \textbf{Realistic, Low-Cost Environment.} To bridge the gap between simulation and real-world driving, we construct a realistic human-in-the-loop driving environment. A fixed 2D top-down navigation map and rear-view mirrors enhance situational awareness, while a Logitech G29 steering wheel and pedal set provide authentic control dynamics and haptic feedback. Together, these create an immersive and realistic setup that effectively captures human driving behavior.
% To narrow the gap between simulation and real-world driving, we design a realistic human-in-the-loop driving environment that closely mimics natural vehicle operation.  
% Specifically, the navigation interface is located on the right side of the main view as a fixed 2D top-down map, imitating the position of in-vehicle navigation screens used in real cars.  
% Rear-view mirrors are also displayed to enhance situational awareness and provide realistic visual feedback.  
% In addition, we use the \textit{Logitech G29 }steering wheel and pedal set as external control devices, allowing participants to perform steering, acceleration, and braking with realistic haptic feedback, thereby ensuring realistic control dynamics and reducing the sim-to-real gap. 
% Together, these components form an immersive and authentic driving environment that effectively bridges simulation and real-world human behavior.
2) \textbf{Diverse Contexts and Multi-modal Data.} Human driving styles vary across environments. To capture diverse behaviors, our system offers multiple CARLA towns, configurable routes, and adjustable traffic densities, alongside events such as intersections, merges, and lane changes to elicit context-dependent driving patterns. Synchronized multi-modal sensor data, including RGB, LiDAR, radar, and semantic segmentation, is recorded under consistent conditions to ensure reproducibility. This environmental and sensory diversity enables comprehensive modeling and systematic analysis of personalized driving behaviors.
% Human driving styles vary across different environments. To capture a broad spectrum of human driving behaviors, our system establishes a highly diverse simulation environment. It supports multiple CARLA towns, varied route configurations, and adjustable traffic densities, enabling data collection under distinct contextual conditions. Moreover, the inclusion of event types such as intersections, merges, and lane changes elicits context-dependent behavioral patterns. The system records synchronized multi-modal sensor data—including RGB, LiDAR, radar, and semantic segmentation—under consistent conditions, ensuring reproducible data acquisition. This integration of environmental and sensory diversity facilitates comprehensive modeling of personalized driving styles and enables systematic analysis of human behavior across a wide range of driving scenarios. 
3) \textbf{Style-Driven Route Design.} To effectively capture personalized driving behaviors, route design must balance duration and scenario diversity. Unlike existing datasets that focus on short event-centric routes or long evaluation routes \cite{carla2023leaderboard}, we design routes of approximately 1 km, providing sufficient temporal and spatial context for style expression while keeping driver workload manageable. All routes are encoded in a standardized XML format with town and configuration metadata, ensuring consistency and reproducibility across drivers and CARLA towns, forming a unified human-in-the-loop data collection framework.
4) \textbf{High Extensibility.} Most importantly, the system is highly extensible, allowing data to be flexibly scaled through this lightweight collection system. Identity labeling, multi-modal synchronization, and style indicator extraction are fully automated, such that drivers only need to operate the vehicle, with no additional manual intervention. This automated pipeline supports efficient scaling to additional drivers and repeated sessions.

\begin{table*}[ht]
\centering
\vspace{-3ex}
\caption{Datasets comparison. \textbf{\#Subjects} is the number of drivers. We give \textbf{Personal Multiple Trajectories} on the same route to eliminate randomness of personal driving behavior. 
``\textbf{--}'' in Subjects indicates unavailable driver identities: StyleDrive lacks driver identity annotations and Bench2Drive contains agent-generated trajectories.} 

% \#Subjects is marked as ``\textbf{--}'' when driver identities are unavailable: StyleDrive lacks driver identity annotations and Bench2Drive contains agent-generated trajectories.

\vspace{-2ex}
% \textbf{Extensibility} denotes that data can be flexibly scaled up by a lightweight capture platform.

\resizebox{\textwidth}{!}{% % ! 表示高度按比例自动调整 
\begin{tabular}{l|c|c|c|c|c|cc}
\toprule
Dataset & Evaluation Type & E2E & \#Subjects & Human Driving & Individual Annotation & Personal Multiple Trajectories  \\ \midrule
HITL-RM \cite{li2023personalized} & Open-Loop & \textcolor{Red}{\ding{55}} & 2 & \textcolor{MyGreen}{\ding{51}} & \textcolor{MyGreen}{\ding{51}} & \textcolor{Red}{\ding{55}}  \\
HITL-CF \cite{zhao2022personalized} & Open-Loop & \textcolor{Red}{\ding{55}} & 4 & \textcolor{MyGreen}{\ding{51}} & \textcolor{MyGreen}{\ding{51}}& \textcolor{Red}{\ding{55}}  \\
HITL-LC \cite{liao2023driver} & Open-Loop & \textcolor{Red}{\ding{55}} & 2 & \textcolor{MyGreen}{\ding{51}} &\textcolor{MyGreen}{\ding{51}} & \textcolor{Red}{\ding{55}}  \\
HITL-Mul \cite{ke2024d2e} & Open-Loop & \textcolor{Red}{\ding{55}} & 60 & \textcolor{MyGreen}{\ding{51}} & \textcolor{MyGreen}{\ding{51}} & \textcolor{Red}{\ding{55}}  \\
UAH \cite{romera2016need} & Open-Loop & \textcolor{Red}{\ding{55}} & 6 & \textcolor{MyGreen}{\ding{51}} &\textcolor{MyGreen}{\ding{51}}& \textcolor{Red}{\ding{55}}  \\
Brain4Cars \cite{jain2016brain4cars} & Open-Loop & \textcolor{Red}{\ding{55}} & 10 & \textcolor{MyGreen}{\ding{51}} &\textcolor{MyGreen}{\ding{51}} & \textcolor{Red}{\ding{55}}  \\
PDB \cite{wei2025pdb} & Open-Loop & \textcolor{Red}{\ding{55}} & 12 & \textcolor{MyGreen}{\ding{51}} & \textcolor{MyGreen}{\ding{51}} & \textcolor{Red}{\ding{55}}  \\
\midrule
StyleDrive \cite{hao2025styledrive} & Semi-Closed-Loop & \textcolor{MyGreen}{\ding{51}} & - & \textcolor{MyGreen}{\ding{51}} & \textcolor{Red}{\ding{55}} & \textcolor{Red}{\ding{55}}  \\
Bench2Drive \cite{jia2024bench2drive} & Closed-Loop & \textcolor{MyGreen}{\ding{51}} & - & \textcolor{Red}{\ding{55}} & \textcolor{Red}{\ding{55}} & \textcolor{Red}{\ding{55}}  \\
Person2Drive (ours) & Closed-Loop & \textcolor{MyGreen}{\ding{51}} & 50 & \textcolor{MyGreen}{\ding{51}} & \textcolor{MyGreen}{\ding{51}} & \textcolor{MyGreen}{\ding{51}}  \\ \bottomrule
\end{tabular}
}
%: Evaluation Type, E2E (End-to-End Capability), Human Driving, Individual Style (whether individual style annotations are made), Subjects (The number of subjects), Repeated sampling, and Extensibility (whether the dataset is scalable).}
\label{com:Dataset}
\vspace{-5ex}
\end{table*}
%%%%%%%

% \begin{figure}
%     \centering
%     \includegraphics[width=1\linewidth]{Fig/style_indices_plot.eps}
%     \caption{Average style indices differences among all individuals.}
%     \label{fig:asid}
% \end{figure}

%%%%%%%

\noindent\textbf{Personalized Dataset}
Based on the proposed data collection system, we introduce \textbf{Person2Drive}, a large-scale and extensible dataset designed to capture diverse human driving behaviors. The dataset supports closed-loop E2E-AD and includes driver identity annotations, making it suitable for individualized driving studies. Unlike existing datasets that primarily emphasize scene diversity, Person2Drive intentionally records multiple trajectories of the same driver on identical routes. Furthermore, the dataset can be expanded using our collection system to capture additional driver styles. 

% In practice, driving a single one-kilometer route requires only about 15 minutes per participant. 
% All identity labeling, multi-modal synchronization, and style indicator extraction 
% are fully automated within the closed-loop system, eliminating the need for 
% additional manual annotation or post-processing. 
% This lightweight and automated pipeline enables efficient expansion of the dataset 
% to more drivers and repeated sessions.
% We collected 30 repetitions of one-kilometer routes from 30 drivers 

% We collected driving data from 50 drivers, each repeatedly driving multiple identical one-kilometer routes with multi-modal sensors, enabling the analysis of both intra-driver consistency and inter-driver style variations, crucial factors for personalized E2E-AD.

We collected driving data from 50 drivers, each repeatedly driving four identical one-kilometer routes with multi-modal sensors, resulting in 55 hours of driving data, 2.3 million LiDAR frames and 16.1 million synchronized camera images. 
The first two routes were repeated eight times and the remaining two routes twice to support both model training and cross-scene evaluation. Moreover, by repeatedly collecting data from the same drivers on identical routes, the dataset enables the analysis of both intra-driver consistency and inter-driver style variations, which are crucial for personalized E2E-AD.
The routes span diverse and geometrically complex scenarios, including signalized and unsignalized intersections, roundabouts, lane reductions, curved merging and diverging structures, highway off-ramps, long gentle curves, and varying lane configurations from single-lane to multi-lane roads. This structured design ensures stylistic diversity while maintaining controlled yet realistic traffic conditions.
% Such structured diversity enables the analysis of both intra-driver consistency and inter-driver style variations under controlled yet realistic traffic conditions.

The dataset offers comprehensive features, including 3D bounding boxes, depth, semantic segmentation, and driver identity labels, all sampled at 10 Hz, serving as a training resource for personalized E2E-AD. To facilitate method reproduction, we adopt a sensor setup similar to Bench2Drive\cite{jia2024bench2drive}, including 1 LiDAR, 6 cameras, 5 radars, 1 IMU \& GNSS, 1 BEV camera, and HD-Map. Compared to existing datasets (Table~\ref{com:Dataset}), Person2Drive captures finer-grained intra- and inter-driver styles, involves more participants, and includes richer multi-modal data, while emphasizing human-in-the-loop driving and personalized driver labeling. More details about the data collection system and dataset can be found in the supplement.

For style annotation, we move beyond predefined categorical labels and instead model each driver as an \textbf{independent style domain}, enabling individual-level personalization. By collecting multiple driving sessions per individual under diverse scenes, our dataset supports fine-grained analysis of intra-driver style changes conditioned on environmental context. Person2Drive explicitly incorporates scene-level semantic annotations that are absent in prior datasets, establishing a comprehensive foundation for benchmarking and advancing research on personalized driving style modeling and fine-grained behavioral understanding.

% \begin{table}[t]
% \centering
% \begin{tabular}{lccc}
% \hline
% Dataset      & Bench2Drive    & PDB & Person2Drive \\ \hline
% Subject     &  Epxert model   & 12  & 20   \\
% Camera      &  3   & 1   & 7    \\
% Depth       &   \textcolor{MyGreen}{\ding{51}}   & \textcolor{Red}{\ding{55}}   & \textcolor{MyGreen}{\ding{51}}   \\
% Semantic    &  \textcolor{MyGreen}{\ding{51}}    &  \textcolor{Red}{\ding{55}}   & \textcolor{MyGreen}{\ding{51}}     \\
% Instance    &  \textcolor{Red}{\ding{55}}   &  \textcolor{Red}{\ding{55}}   & \textcolor{MyGreen}{\ding{51}}      \\
% Scene       & Single  &  Single   & Diverse     \\
% Neighbor Vehicle & \textcolor{Red}{\ding{55}} & \textcolor{Red}{\ding{55}}   & \textcolor{MyGreen}{\ding{51}}      \\ \hline
% \end{tabular}
% \caption{Datasets Comparison.}
% \label{com:Dataset}
% \end{table}

\vspace{-2ex}
\subsection{Driving Style Evaluation}

Evaluating driving style is a critical yet underexplored aspect of personalized E2E-AD. The lack of standardized metrics prevents meaningful evaluation of inter-driver differences and model performance. Therefore, we present a three-step evaluation framework for personalized E2E-AD: 1 constructing style vectors to capture key dimensions of individual driving behavior, 2) quantifying inter-driver differences, and 3) applying scenario-specific normalization for consistent comparison across diverse contexts. Built on this framework, we introduce MMDSS, providing the first reproducible, quantitative analysis of driving styles under the E2E-AD setting.

% Built on this framework, we introduce MMDSS, providing the first reproducible, quantitative analysis of driving styles.

\noindent\textbf{Style Vector Indicators}
To comprehensively capture individual driving styles and highlight inter-driver differences, we design style vector indicators based on vehicle control parameters and driver behavior data. To identify the most distinctive trajectory-derived style indicators, we introduce a median normalization and variance-based discriminability method, inspired by classical robust statistical analysis \cite{moore2009introduction}. 
For each driver, we compute the indicator's mean and normalize it by the population median, capturing deviations from typical behavior. We then measure the variability of these normalized values across all drivers and select the top ten indicators with the highest standard deviation as representative style features, as Eq.~\ref{norm}.
\vspace{-1ex}
\begin{equation}
    x_{std} = \sqrt{\frac{1}{N}\sum_{i=1}^{N}(x'-\mu)^2},
\label{norm}
\end{equation}
where $x'=x-x_{mid}$ and $x_{mid}$ means the median. This process robustly highlights inter-driver differences while mitigating scale and outlier effects. More details can be found in Section~\ref{sec5.1} and Supplementary Material. 

\noindent\textbf{Style Differences Quantification}
To better quantify differences across driving styles, we introduce MMDSS, a distribution-level metric that models driving style as a behavioral distribution rather than isolated trajectory instances. This can capture global style tendencies and provide a more stable measure of inter-style divergence. We also report the KL divergence to further validate the separability between different style distributions.

Specifically, given two sets of trajectory features $X = \{x_i\}_{i=1}^m$ and $Y = \{y_j\}_{j=1}^n$, the empirical MMD is computed as in Eq.~\ref{eq:mmd}:
\vspace{-2mm}
\begin{equation}
\begin{split}
\text{MMD}^2(X, Y) =
& \frac{1}{m^2}\sum_{i=1}^{m}\sum_{i'=1}^{m} k(x_i, x_{i'}) \\
 + \frac{1}{n^2}\sum_{j=1}^{n}\sum_{j'=1}^{n} k(y_j, y_{j'}) 
& - \frac{2}{mn}\sum_{i=1}^{m}\sum_{j=1}^{n} k(x_i, y_j),
\end{split}
\label{eq:mmd}
\end{equation}
where $k(\cdot,\cdot)$ denotes an RBF kernel function here. While MMD quantifies distributional divergence, it is not directly interpretable as a similarity score.
Therefore, we introduce MMDSS, a normalized similarity score defined as:
\vspace{-1ex}
\begin{equation}
    MMDSS = \frac{1}{1+MMD}.
    \label{MMDSS}
\vspace{-1ex}
\end{equation}

MMDSS transforms the raw divergence measure into a bounded, interpretable score in $[0,1)$, where higher values indicate more consistent driving styles. This makes MMDSS particularly suitable for evaluating intra- and inter-individual behavioral similarity in personalized driving.

In addition to MMD, we also employ KL divergence to assess distributional differences in driving styles further. In our implementation, KL is computed using a histogram-based symmetric formulation with 50 bins, independent min--max normalization to $[0,1]$, and $\epsilon$-smoothing ($10^{-8}$). Lower KL value indicates a closer match between the two style distributions, providing a complementary measure of style similarity. 

% The KL divergence from distribution Q to distribution P is given by Eq.~\ref{eq:KL}:
% \begin{equation}
% \begin{split}
% D_{\text{KL}}(P \parallel Q) &= \sum_{x} P(x) \log \frac{P(x)}{Q(x)} \end{split},
% \label{eq:KL}
% \end{equation}
% where P(x) and Q(x) are the probability density functions of distributions P and Q.

By combining both MMD and KL divergence, we can capture a comprehensive understanding of how different drivers' styles diverge not only in terms of distributional distances (MMD) but also in terms of relative entropy (KL divergence). This multi-metric approach offers a robust and interpretable analysis of driving styles.

\noindent\textbf{Scenario-specific Normalization}
In addition, driving behavior is inherently context-dependent. For example, a speed of 80 km/h may be considered aggressive in an urban road segment but conservative on a highway. To account for such context-dependent variations, we adopt a scenario-specific normalization strategy. Here, each route is treated as a distinct scenario. Specifically, we analyze the feature distributions of all drivers within each route and perform min--max normalization separately. This design mitigates style bias induced by route-dependent feature shifts and enables more consistent comparison of individual driving behaviors across heterogeneous driving environments.

% In addition, when people drive in different scenarios, they often exhibit distinct driving habits. To further account for style variations across scenarios, we adopt a scene-aware normalization strategy.
% Specifically, since the same trajectory may represent different styles under different contexts, we analyze the feature distributions of all drivers within each scenario and perform min–max normalization separately for each. This normalization mitigates the style bias caused by scenario-dependent feature shifts and enables a more consistent comparison of individual driving behaviors across diverse conditions.

\vspace{-1ex}

\subsection{Personal Driving Style Adaptation}
Based on the proposed style evaluation model, we propose a Personal Driving Style Adaptation (\textbf{PDSA}) strategy to guide the E2E-AD model to be more similar to the target style while maintaining great performance. Our entire training framework includes three stages: E2E Basic Model training, Driving Style Modeling, and Personal Style-Guided Finetuning, as illustrated in Fig.~\ref{fig:framework}.

% \begin{figure*}
%     \centering
%     \includegraphics[width=1\linewidth]{Fig/method_1.png}
%     \caption{Framework of Personal Driving Style Adaptation.}
%     %We first employ Basic Model which is capable for close-loop autonomous driving task. Then, we design and train a reward model modeling the driving style from trajectories. After obtaining the reward model, we utilize it to finetune the e2e trajectory planner with adding additional trajectory proposal level style difference loss.}
%     \label{fig:method}
%      \vspace{-2ex}
% \end{figure*}

\noindent
\textbf{E2E Basic Model training.} We first employ DiffusionDrive \cite{liao2025diffusiondrive} as our basic driving model, leveraging its ability to capture complex temporal dynamics in vehicle motion. The model is trained using historical trajectory sequences alongside the BEV features extracted from the environment on the Bench2Drive dataset. This combination allows the backbone to effectively encode both past motion patterns and spatial context, providing a strong foundation for basic driving performance and even style-aware refinement.

\noindent
\textbf{Driving Style Modeling.}   
To bridge raw trajectory data and interpretable driving style features, we introduce a learnable reward model that replaces handcrafted style formulas. Given a trajectory, the model outputs a 10-dimensional style vector by learning a differentiable mapping between trajectory dynamics and driver style. This continuous supervision enables generalization to unseen behaviors, which fixed analytic functions cannot achieve. The model is implemented as a lightweight MLP with two hidden layers of 256 units each and ReLU activations. To incorporate environmental context, we encode the BEV map into a global scene vector, inject it at each time step of the trajectory encoder, and aggregate temporal features through the MLP to produce the final style vector.

% To bridge raw trajectory representations and interpretable style features, we introduce a reward model to facilitate PDSA.   
% Rather than directly computing style metrics through handcrafted formulas, we train a learnable mapping that captures the underlying correspondence between trajectory dynamics and driver style.   
% This formulation allows the style supervision to remain differentiable and generalizable to unseen trajectories, which would not be achievable with fixed analytic functions.   
% The reward model takes the trajectory as input and outputs a 10-dimensional style vector representing the driver’s style, corresponding to the ten representative style features introduced above.   
% During training, the predicted style vector of the generated trajectory is aligned with that of the corresponding real trajectory, providing a style supervision signal for subsequent fine-tuning.   
% To further explore the potential contribution of environmental context, we design a variant of the reward model that additionally incorporates BEV features as auxiliary input.

\noindent
\textbf{Personal Style-Guided Finetuning.} Given a reward model that maps driving trajectories to style vectors, we leverage it to fine-tune the end-to-end driving model, updating only the trajectory prediction head. All predicted trajectories are mapped to style vectors via the reward model and aligned with the corresponding ground-truth style vectors using mean squared error. This alignment loss is incorporated into the fine-tuning process, enhancing the trajectory prediction head’s sensitivity to style while avoiding performance degradation that could result from direct trajectory-level supervision. In this way, PDSA bridges style evaluation and policy optimization, enabling style-consistent driving behaviors without compromising baseline driving competence.

% \begin{figure*}
%     \centering
%     \includegraphics[width=1\linewidth]{Fig/method.png}
%     \caption{The pipeline of Style-Guided Fine-tuning with 3 stages, including E2E base model preparation, Driving style modeling, and driving style guided fine-tuning.}
%     \label{fig:method}
% \end{figure*}

\vspace{-2ex}
\section{Experiment}
\vspace{-2ex}
In this section, we conduct extensive experiments on both our proposed dataset and the StyleDrive dataset to evaluate the effectiveness of the proposed style  assessment system and the performance of the proposed fine-tuning framework. 

\vspace{-2ex}

\subsection{Driving Style Evaluation}
\vspace{-2ex}

% \begin{wrapfigure}{r}{0.6\linewidth}  % l=左，r=右，宽度0.45列
% \vspace{-5ex}
%     \centering
%     \includegraphics[width=\linewidth]{Fig/bar_chart2.eps}
%     \vspace{-4ex}
%     \caption{Style Modeling results on the Person2Drive dataset.}
%     \label{fig:rmp2d0}
% \vspace{-4ex} 
% \end{wrapfigure}

\begin{figure}[!ht]
\vspace{-3.5ex}
\centering
\begin{subfigure}{0.585\linewidth}
    \centering
    \includegraphics[width=\linewidth]{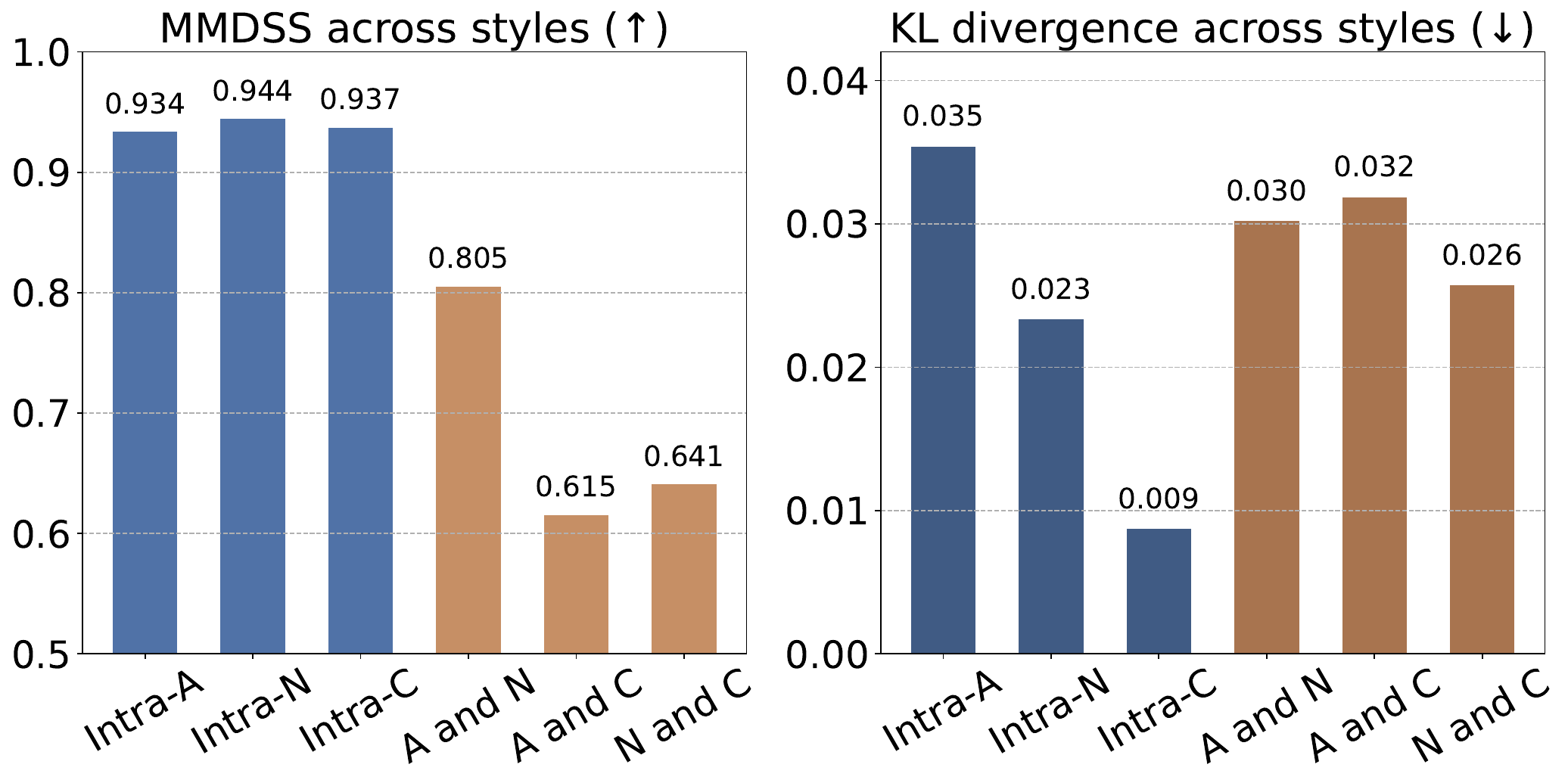}
    \caption{StyleDrive Dataset}
    \label{fig:1a}
\end{subfigure}
\hspace{0.02\linewidth}
\begin{subfigure}{0.316\linewidth}
% \vspace{-0.4ex}
    \centering
    \includegraphics[width=\linewidth]{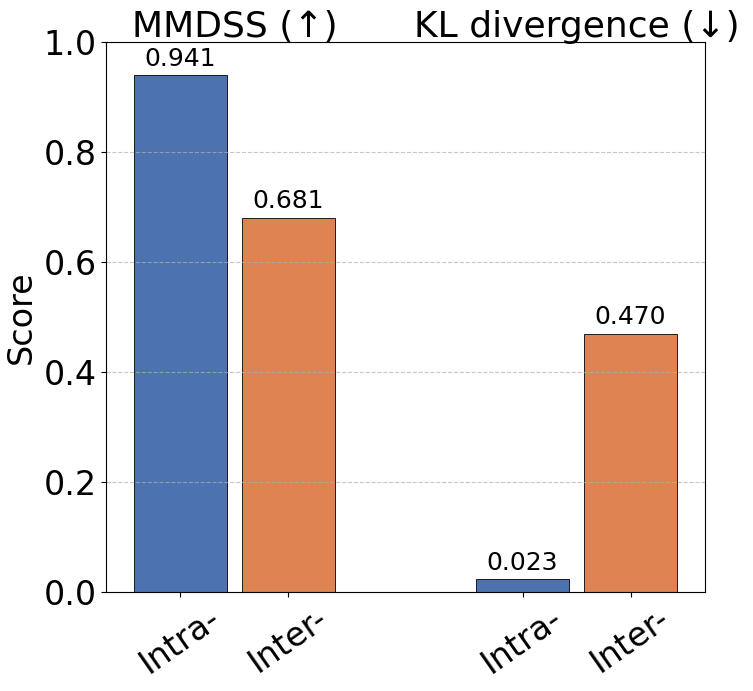}
    \caption{Person2Drive Dataset}
    \label{fig:1b}
\end{subfigure}
\vspace{-1.5ex}
\caption{Style similarity evaluation results on the GT trajectories of StyleDrive dataset and Person2Drive dataset.(a) Results on the StyleDrive dataset. (b) Results on the Person2Drive dataset.}
\label{fig:sefbar}
\vspace{-6ex}
\end{figure}

We first validate the proposed style evaluation framework. A reasonable and effective style evaluation system should ensure that intra-style differences are smaller than inter-style differences. Therefore, on the StyleDrive dataset, we compute style vectors from the ground-truth trajectories and calculate MMDSS and KL divergence across different style categories. By comparing intra-style and inter-style differences, we verify the effectiveness of our driving style evaluation, as shown in 
Figure~\ref{fig:sefbar}(a). 

It presents MMDSS and KL divergence across intra-style (Intra-A/N/C) and inter-style comparisons. MMDSS behaves as expected—values within the same style remain high (about 0.93), while cross-style values drop notably (0.61–0.80), confirming its reliability in capturing style consistency. In contrast, KL divergence shows partial deviation: although Intra-N and Intra-C exhibit small intra-style KL values, Intra-A is unexpectedly higher. This irregularity may arise from distributional imbalance or estimation instability, highlighting KL’s sensitivity to sample support and noise. Overall, our proposed MMDSS proves to be a robust indicator of intra-style similarity. 

% \begin{figure}[h]
%     \centering
% \includegraphics[width=0.5\linewidth]{Fig/bar_chart.eps}
%     \vspace{-2mm}
%     \caption{Style similarity evaluation results on the GT trajectories of the StyleDrive dataset, which reports average results among all styles.}
%     \label{fig:sefbar}
%      \vspace{-3ex}
% \end{figure}

To further examine whether the proposed driving style evaluation framework can capture fine-grained individual style differences, we conduct additional experiments on the Person2Drive dataset. Specifically, based on the extracted style vector indicators, we first compute MMDSS and KL divergence across multiple trajectories of each driver to quantify intra-individual style consistency. We then compute the same metrics between different drivers to assess inter-individual style differences, thereby validating the framework’s capability to effectively distinguish personalized driving behaviors. 

% \begin{wrapfigure}{r}{0.35\linewidth}  % l=左，r=右，宽度0.45列
% \vspace{-8ex}
%     \centering
%     \includegraphics[width=\linewidth]{Fig/driver_comparisonv1.eps}
%     \vspace{-9ex}
%     \caption{Style Modeling results on the Person2Drive dataset.}
%     \label{fig:rmp2d2}
% \vspace{-8ex} 
% \end{wrapfigure}

As shown in Figure~\ref{fig:sefbar}(b), intra-person MMDSS (0.941) exceeds inter-person MMDSS (0.681), and intra-person KL (0.023) is much lower than inter-person KL (0.470), demonstrating strong within-driver consistency and clear between-driver separability. These results confirm that our evaluation framework reliably captures intra- and inter-person driving style differences and provides an interpretable, quantitative basis for analyzing personalized driving styles.
\vspace{-1ex}

% \begin{table}[ht]
%     \centering
%     \caption{Results of the proposed style evaluation framework on the Person2Drive dataset. Both MMDSS and KL can effectively capture consistent and discriminative personalized driving styles.}
%     \vspace{-2ex}
%     \scalebox{0.9}{
%     \begin{tabular}{c|ccc}
%     \toprule
%          &  Intra-Person & Inter-Person & Abs. $\Delta$ \\ 
%          \midrule
%        MMDSS  (↑)  &  0.9267  & 0.6545 & 0.2722 \\
%        KL (↓) &  0.0244  &  0.5119 & 0.4875 \\ \hline
%     \end{tabular}
%     }
%     \label{tab:resP2D}
% \end{table}

% As shown in Table \ref{tab:resP2D}, our proposed style evaluation framework effectively captures both intra- and inter-personal driving style differences. The intra-person MMDSS (0.9267) is substantially higher than the inter-person MMDSS (0.6545), demonstrating strong style consistency within individual drivers and clear separability across different ones. Similarly, the intra-person KL divergence (0.0244) is significantly lower than the inter-person KL (0.5119), further validating that our metrics can reflect fine-grained behavioral distinctions at the individual level. Together, these results indicate that the proposed framework provides a reliable and interpretable quantitative foundation for personalized driving style analysis.

\vspace{-2ex}
\subsection{Reward Model-based Style Modeling}
% \vspace{-1ex}
We evaluate the proposed reward model-based style modeling framework on the StyleDrive dataset by comparing the style discrepancy metrics of the style vectors predicted by our reward model (\textbf{Ours}) with those of the ground-truth vectors (\textbf{GT}), considering both intra-style and inter-style differences. Scenario-level segmentation is also performed to assess the model’s ability to capture fine-grained style variations. The results, measured by MMDSS and KL divergence (Table~\ref{tab:rm}), show that the learned style vectors closely match the ground-truth representations.
Intra- and inter-style MMDSS indicate that the learned style vectors align well with the real styles, with minor deviations in certain scenarios. Both MMDSS and KL divergence confirm that intra-style differences are smaller than inter-style differences within the same scenario. These results demonstrate that the reward model can effectively learn fine-grained personalized driving styles from user trajectories, with MMDSS providing a robust and interpretable measure of style consistency and separability.
\vspace{-3ex}
\begin{table*}[ht]
\small
\caption{Reward model-based style modeling results on the StyleDrive dataset. The table shows the differences ($\Delta$) between the learned styles and the ground-truth across various scenarios, illustrates that the learned representations are highly consistent with the real ones in reflecting both intra-style and inter-style variations. }

% The figure shows the differences ($\Delta$) between the learned styles and the ground-truth across various scenarios to evaluate whether the proposed reward model effectively captures style representations. The results show that the learned representations are highly consistent with the real ones in reflecting both intra-style and inter-style variations.
\vspace{-1ex}
\resizebox{\textwidth}{!}{% % ! 表示高度按比例自动调整
\begin{tabular}{c|c|ccc|ccc|ccc}
\hline
                              & Scene     & \multicolumn{3}{c|}{Lane Following}              & \multicolumn{3}{c|}{Special Interior Road}       & \multicolumn{3}{c}{Protected Intersections}      \\ \cline{2-11} 
                              & Model     & GT     & Ours   & $\Delta$                       & GT     & Ours   & $\Delta$                       & GT     & Ours   & $\Delta$                       \\ \hline
                              & MMDSS (↑) & 0.9427 & 0.9511 & {\color[HTML]{FE0000} 0.0084}  & 0.9548 & 0.9575 & {\color[HTML]{FE0000} 0.0027}  & 0.9175 & 0.9129 & {\color[HTML]{FE0000} -0.0046} \\
\multirow{-2}{*}{Intra-Style} & KL (↓)    & 0.0194 & 0.0171 & {\color[HTML]{FE0000} -0.0023} & 0.0157 & 0.0083 & {\color[HTML]{FE0000} -0.0074} & 0.0323 & 0.0263 & {\color[HTML]{FE0000} -0.0060} \\ \hline
                              & MMDSS (↑) & 0.6771 & 0.6783 & {\color[HTML]{FE0000} 0.0012}  & 0.6790 & 0.6762 & {\color[HTML]{FE0000} -0.0028} & 0.7052 & 0.7072 & {\color[HTML]{FE0000} 0.0020}  \\
\multirow{-2}{*}{Inter-Style} & KL (↓)    & 0.0272 & 0.0233 & {\color[HTML]{FE0000} -0.0039} & 0.0187 & 0.0109 & {\color[HTML]{FE0000} -0.0078} & 0.0419 & 0.0341 & {\color[HTML]{FE0000} -0.0078} \\ \hline
\end{tabular}
}
\label{tab:rm}
\vspace{-4ex}
\end{table*}

\begin{wrapfigure}{r}{0.55\linewidth}  % l=左，r=右，宽度0.45列
    \vspace{-6ex}
    \centering
    \includegraphics[width=\linewidth]{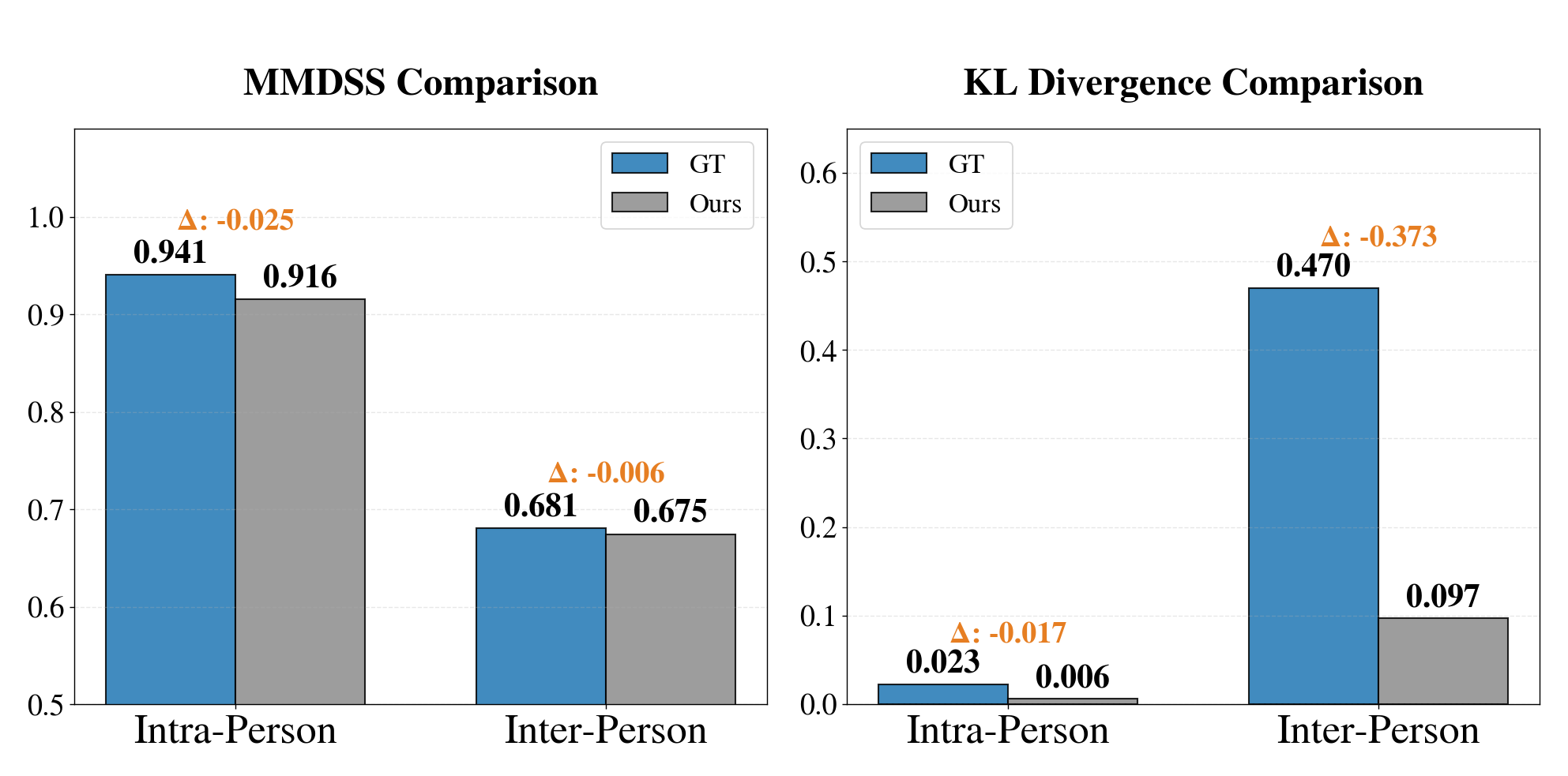}
    \vspace{-3ex}
    \caption{Style Modeling results on the Person2Drive dataset.}
    \label{fig:rmp2d}
\vspace{-6ex} 
\end{wrapfigure}

We then also conduct a similar analysis on the Person2Drive dataset to verify the effect of our method. Specifically, we compare the ground-truth and reward-model outputs at both intra- and inter-person levels to validate whether the reward model can effectively capture and represent driving styles across individuals. 
As illustrated in Figure~\ref{fig:rmp2d}, the reward-based style representations closely align with the ground-truth at both intra- and inter-person levels. MMDSS effectively captures the expected pattern, with smaller intra-person differences and larger inter-person differences, whereas KL values are notably lower, particularly for inter-person comparisons, due to their sensitivity to compact distributions. By directly comparing sample distributions, MMDSS provides a more robust and interpretable measure of style consistency and separability, confirming that the reward model reliably learns fine-grained, individualized driving styles.
% \usepackage{multirow}

% \begin{table}[t]
%     \centering
%     \begin{tabular}{c|c|c|c}
%     \toprule
%      \textbf{Scene Name} & \textbf{A and N} & \textbf{N and C} & \textbf{C and A} \\
%     \midrule
%     \multicolumn{4}{c}{\textbf{With BEV}} \\
%     \midrule
%      lane\_following          & 0.8141 & 0.6301 & 0.5835 \\
%      special\_interior\_road   & 0.8013 & 0.6383 & 0.6010 \\
%      protected\_intersections  & 0.8220 & 0.6362 & 0.6663 \\
%     \midrule
%     \multicolumn{4}{c}{\textbf{W/O BEV}} \\
%     \midrule
%      lane\_following          & 0.9047 & 0.6398 & 0.6211 \\
%      special\_interior\_road   & 0.7822 & 0.6179 & 0.5794 \\
%      protected\_intersections  & 0.8585 & 0.6664 & 0.7072 \\
%     \bottomrule
%     \end{tabular}
%     \caption{MMD Similarity Scores for Different Scenes (With BEV vs. W/O BEV)}
%     \label{tab:rm2}
% \end{table}

\vspace{-2ex}
\subsection{Personal Driving Style Adaptation Results}
\vspace{-1ex}
To verify the performance of the proposed PDSA framework, we first conducted preliminary experiments on the StyleDrive dataset, followed by formal experiments on the Person2Drive dataset. 
Specifically, for the StyleDrive, we fine-tune the baseline  using aggressive driving data, while for the Person2Drive, we adopt each individual's historical driving trajectory to personalize the model. Then we compare the MMDSS between the style indicators before and after fine-tuning. We also compare different variants of our proposed PDSA as ablation studies, including: \textbf{DFT}, which means directly fine-tuning; \textbf{PDSA-WB}, which means using reward models without BEV features; and \textbf{PDSA}, which means using reward models with BEV features. Results are shown in Tables~\ref{tab:navsim_ft} and \ref{tab:ftp_sig_merged}.

% We report the average performance on all subjects under different scenarios in Table~\ref{tab:ftp_sig_merged}.  

Table~\ref{tab:navsim_ft} reports the MMDSS results for the generated trajectories after fine-tuning with the aggressive driving subset. All three fine-tuning strategies progressively shift the model toward the aggressive style, with the MMDSS between the aggressive style (A) and the ground truth increasing from 0.8453 to 0.8605. Among the methods, PDSA achieves the largest improvement (+0.0152), indicating its superior capability in capturing the desired stylistic characteristics.

\begin{table*}[ht]
\vspace{-3ex}
\centering
\caption{Evaluation of reward model training and fine-tuning with Aggressive driving dataset based on the driving style labels (\textbf{A}ggressive - \textbf{N}ormal - \textbf{C}onservative) of StyleDrive. }
\vspace{-2ex}
\begin{tabular}{lcccccc}
\hline
Method       & A to GT& $\Delta$ & N to GT& $\Delta$ & C to GT & $\Delta$ \\ \hline
baseline&   0.8453& 0 &  0.9439& 0 &  0.7158 & 0\\
DFT& 0.8551 & +0.0098 &   0.9327 & -0.0112 &   0.7103 & -0.0055\\
PDSA-WB&    0.8601& +0.0148 &  0.9290&  -0.0149 &    0.7080 &  -0.0078\\
\textbf{PDSA} &    \textbf{0.8605}&  \textbf{+0.0152} & 0.9308  & -0.0131 &   0.7088 & -0.0070 \\
\hline
\end{tabular}
\label{tab:navsim_ft}
\vspace{-4ex}
\end{table*}

Table~\ref{tab:ftp_sig_merged} reports the performance gains across both scenarios after applying the PDSA framework. DFT yields modest gains over the baseline, indicating personal driving histories provide useful style cues. Incorporating the reward model without BEV features further amplifies these improvements, demonstrating the benefits of style-guided supervision. Our full model, PDSA, achieves the best performance in both scenarios, showing that adding BEV context enables more accurate style alignment. In addition, per-driver gains $\Delta$ (method$-$baseline) indicate that most drivers benefit, with all improvements statistically significant according to \textbf{Wilcoxon signed-rank tests (p<0.001)}. Overall, these results validate that PDSA effectively injects personalized style into the trajectory prediction head while maintaining stable gains across diverse driving scenarios.

% We further conduct an additional comparison between direct style-vector supervision and the proposed reward model using a subset of drivers. The reward model achieves a higher MMDSS score (0.7055 vs. 0.6690), indicating more accurate style alignment than direct style-vector supervision.

\begin{table}[ht]
\vspace{-3ex}
\centering
\small
\setlength{\tabcolsep}{2.5pt}
\renewcommand{\arraystretch}{1.1}
\caption{\textbf{Personalization results on Person2Drive.}
The table presents MMDSS and per-driver robustness statistics
($\Delta=$ method$-$baseline, multiplied by $10^{-3}$).}
\vspace{-2ex}
\label{tab:ftp_sig_merged}

\resizebox{\textwidth}{!}{
\begin{tabular}{l|ccc|ccc|c}
\toprule
\multirow{2}{*}{\textbf{Method}}
& \multicolumn{3}{c|}{\textbf{Route 1}}
& \multicolumn{3}{c|}{\textbf{Route 2}}
& \multirow{2}{*}{\textbf{MMDSS Avg$\uparrow$}} \\
\cline{2-7}
& \textbf{MMDSS$\uparrow$}
& \textbf{$\Delta$Mean$\pm$Var}
& \textbf{\%($\Delta>0$)$\uparrow$}
& \textbf{MMDSS$\uparrow$}
& \textbf{$\Delta$Mean$\pm$Var}
& \textbf{\%($\Delta>0$)$\uparrow$}
&  \\
\midrule

Baseline
& 0.6410 & \textbf{--} & \textbf{--}
& 0.6894 & \textbf{--} & \textbf{--}
& 0.6652 \\

DFT
& 0.6425 & $1.597 \pm 1.009$ & 82\%
& 0.7095 & $20.163 \pm 0.845$ & 70\%
& 0.6760 \\

PDSA-WB
& 0.6439 & $2.945 \pm 0.467$ & 76\%
& 0.7131 & $23.720 \pm 1.094$ & 58\%
& 0.6785 \\

\textbf{PDSA}
& \textbf{0.6465} & $\mathbf{5.590} \pm \mathbf{0.225}$ & \textbf{84\%}
& \textbf{0.7135} & $\mathbf{24.151} \pm \mathbf{0.948}$ & \textbf{76\%}
& \textbf{0.6800} \\

\bottomrule
\end{tabular}
}
\vspace{-5ex}
\end{table}

\vspace{-2ex}
\subsection{Comparison with coarse style categories (ANC).}
% \vspace{-0.5ex}

\begin{wraptable}[4]{r}{0.5\linewidth}  % l=左，表格占宽度45%
\vspace{-5ex}
\centering
% \small
\normalsize
\setlength{\tabcolsep}{3pt}
\renewcommand{\arraystretch}{0.7}
\vspace{-3ex}
\caption{ANC coarse-style fine-tuning vs. personalized fine-tuning on Person2Drive.}
\resizebox{\linewidth}{!}{%
\begin{tabular}{lcccc}
\toprule
\textbf{Method} & \textbf{A} & \textbf{N} & \textbf{C} & \textbf{Personal (ours)} \\
\midrule
\textbf{Average} & 0.6156 & 0.6125 & 0.6174 & 0.6800 \\
\bottomrule
\end{tabular}%
}

\label{tab:anc_ft}
\vspace{-5ex}
\end{wraptable}

To further compare with the prior coarse-grained style supervision (e.g., StyleDrive), 
we build three category-level fine-tuning baselines by training the base model on trajectories from each ANC group (Aggressive/Normal/Conservative), resulting in three ANC-specific models.
For each driver, we select the ANC-specific model whose driving style is most similar to the driver and compute their style similarity. We then compare this category-level baseline with fine-tuning on the driver’s own trajectories.
As shown in Table~\ref{tab:anc_ft}, ANC-specific model yield weaker style alignment than fine-tuning on individual trajectories,
while our personalized approach enables the model to better align its driving style with that of the driver.
Overall, this comparison highlights the limitations of coarse category-level supervision and demonstrates the effectiveness of personalized fine-tuning.

\subsection{Additional Validation}

% \noindent\textbf{Reward Model Validation.}
% To validate the effectiveness of the reward model, we compare it with direct style-vector supervision using a subset of drivers. The reward model achieves a higher MMDSS score (0.7055 vs. 0.6690), indicating more accurate style alignment than direct style-vector supervision.

\begin{wraptable}[4]{r}{0.56\columnwidth}
\vspace{-3.5em}
\centering
\scriptsize
\caption{\textbf{Alternative Formulations}}
\label{tab:alternative_formulations}
\resizebox{0.55\columnwidth}{!}{
\begin{tabular}{lccccc}
\toprule
 & DSV & BC & IRL & PL & PDSA \\
\midrule
MMDSS & 0.6690 & 0.6811 & 0.6591 & 0.6796 & \textbf{0.7055} \\
\bottomrule
\end{tabular}
}

\end{wraptable}
\textbf{Alternative Formulations.} 
We compare PDSA with several alternative style-learning formulations, including direct style-vector supervision(DSV), behavior cloning (BC), inverse reinforcement learning (IRL), and preference learning(PL). As shown in Table~\ref{tab:alternative_formulations}, PDSA achieves the highest MMDSS among all compared methods, indicating more accurate style alignment than alternative formulations.

% preamble 里需要

% 正文中：把 wraptable 放在你希望它出现的段落前
\begin{wraptable}[4]{r}{0.52\columnwidth}
\vspace{-3.2em}
\centering
\setlength{\tabcolsep}{3pt}
\renewcommand{\arraystretch}{0.52}
\caption{\textbf{Human Preference Study}}
\resizebox{\linewidth}{!}{
\begin{tabular}{lccc}
\toprule
Method & Route 1 (\%) $\uparrow$ & Route 2 (\%) $\uparrow$ & Participants \\
\midrule
Baseline & 24.0 & 28.0 & 50 \\
PDSA (Ours) & \textbf{76.0} & \textbf{72.0} & 50 \\
\bottomrule
\end{tabular}
}
\vspace{-0.6em}

\label{tab:user_study}
\vspace{-3.0em}
\end{wraptable}
\noindent\textbf{Human Preference Study.}
We conduct a human preference study involving 50 participants. As shown in Table~\ref{tab:user_study}, PDSA is consistently preferred over the baseline in both scenes, suggesting that the improvements measured by MMDSS align well with human perception of driving style.

\noindent\textbf{Driver-Specific Behavior Analysis.}
To distinguish driver-specific behavior from scenario-dependent effects, we compare trajectories generated by different personalized models under identical scenes. The inter-driver MMDSS scores are 0.6383 and 0.6716 in Route~1 and Route~2, respectively, which are lower than the corresponding baseline-personal trajectory similarities (0.6410 and 0.6894),indicating that the learned policies capture driver-specific behavioral characteristics rather than converging to a common scenario-dependent policy.

% \begin{table}[ht]
% \centering
% \small
% \setlength{\tabcolsep}{4pt}
% \renewcommand{\arraystretch}{1.05}
% \caption{ANC coarse-style fine-tuning vs. personalized fine-tuning on Person2Drive.}
% \vspace{-1ex}
% \label{tab:anc_ft}
% \begin{tabular}{l|ccc}
% \toprule
% \textbf{Method} & \textbf{Scene 1} & \textbf{Scene 2} & \textbf{Average} \\
% \midrule
% ANC-A (matched) & -- & -- & -- \\
% ANC-N (matched) & -- & -- & -- \\
% ANC-C (matched) & -- & -- & -- \\
% \midrule
% Personal (ours) & -- & -- & -- \\
% \bottomrule
% \end{tabular}
% \end{table}

\section{Analysis and Discussion}
\vspace{-1ex}
\subsection{Driving Style Indicator Analysis}
\begin{wrapfigure}{r}{0.53\linewidth}  % l=左，r=右，宽度0.45列
% \vspace{-1ex}  % 可以微调与上一段文字的间距
\centering
\includegraphics[width=\linewidth]{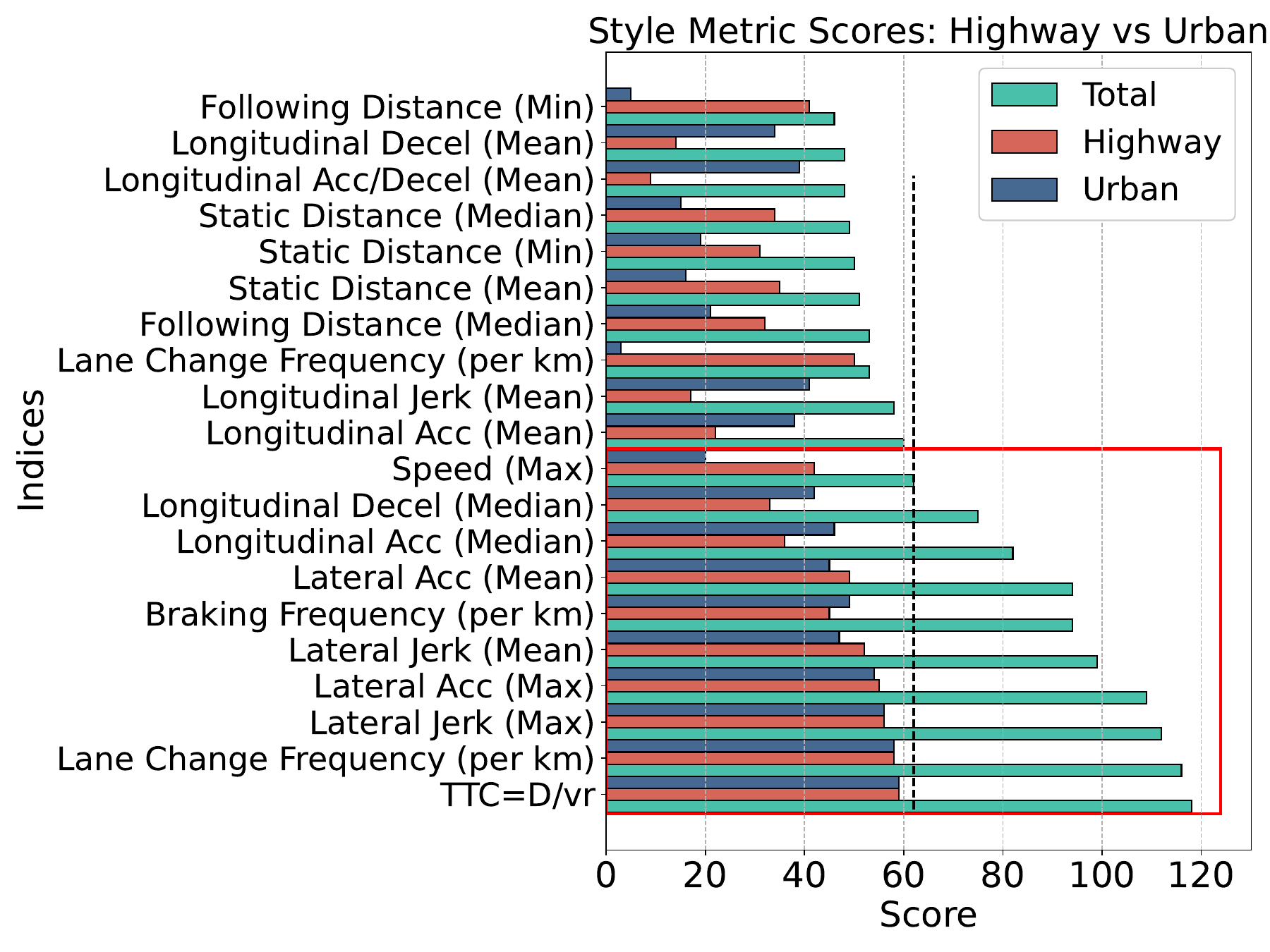}
\vspace{-5.4ex}  % 可以微调图和 caption 之间的间距
\caption{Average style indices differences among all individuals.}
\label{fig:asid}
\vspace{-5ex}  % 调整图底部与下一行文字间距
\end{wrapfigure}

In this section, we analyze the effectiveness of the proposed style indices. Specifically, during the collection of the Person2Drive dataset, we record the corresponding potential style indicators. The measured differences across drivers for the top 20 most discriminative indicators are presented in Figure~\ref{fig:asid}, demonstrating the ability of these indices to capture inter-driver variability.
The figure reveals that the scores capture the divergence of driving behaviors across different scenarios. Metrics such as TTC and Lane Change Frequency (per km) 
\label{sec5.1}
achieve higher scores on highways, indicating greater variability among individual driving trajectories in these metrics. Urban scenarios show elevated scores for metrics like Longitudinal Acceleration, reflecting pronounced differences in stop-and-go behavior across drivers.
The total scores highlight the most influential metrics, with the tenth-ranked metric annotated to delineate high-impact indicators from secondary ones. Overall, the grouped bar plot conveys each metric's relative contribution to driving style variability, providing quantitative guidance for driving style analysis and E2E-AD personalization.

% \begin{figure}[t]
%     \centering
% \includegraphics[width=0.7\linewidth]{Fig/style_indices_plot_horizontal.eps}
%   % \vspace{-3ex}
%     \caption{Average style indices differences among all individuals.}
%     \label{fig:asid}
%     % \vspace{-5ex}
% \end{figure}

\vspace{-1ex}

\subsection{Does style-guided fine-tuning affect driving ability?}
\vspace{-1ex}
The style-guided fine-tuning enables the model to align the generated trajectory with the real driving style. However, it remains unclear whether such stylized fine-tuning affects driving ability. Here, we compare the basic driving ability before and after PDSA in terms of driving score and success rate. We further extend this comparison to diverse and difficult cases in Bench2Drive to evaluate performance under \textbf{challenging scenarios}. 
As shown in Table~\ref{tab:driving_ability_side}, style-guided fine-tuning (PDSA) improves driving performance across datasets. On the Person2Drive dataset, the driving score increases by 11.91 points, with only a negligible decrease of 0.5\% in success rate. On the Bench2Drive dataset, the driving score increases by 0.78 points and success rate improved by 2.18\%. 
These results suggest that aligning trajectories with personalized styles does not degrade driving performance, demonstrating that our fine-tuning approach effectively balances style adaptation with fundamental driving ability.

% These results indicate that incorporating individual driving style not only preserves but can even enhance the model’s driving performance, demonstrating that style-guided fine-tuning consistently maintains or slightly improves driving outcomes while injecting personalized style into the trajectory prediction head.

% \begin{table}[ht]
% \small
% \centering
% \caption{Driving ability comparison among all individuals before and after personal style-guided finetuning}
% \vspace{-2ex}
% \scalebox{0.875}{
% \begin{tabular}{l|ccc}
% \toprule
%  & Driving Score $\uparrow$ & Success rate (\%) $\uparrow$ \\
% \midrule
% GT & 100 & 1.0 \\
% Original & 58.22 & 0.989 \\
% Fine-tuning & 72.78 & 0.982 \\
% $\Delta$ & +14.56 & -0.007 \\
% \hline
% \end{tabular}
% }
% \label{tab:cd}
% \vspace{-2ex}
% \end{table}

% \begin{table}[ht]
% \centering
% \caption{Comparison of driving performance before and after style-guided fine-tuning (PDSA). Higher values indicate better performance.}
% \label{tab:cd2}
% \begin{tabular}{lcc}
% \toprule
% \textbf{Method} & \textbf{Driving Score$\uparrow$} & \textbf{Success Rate$\uparrow$} \\
% \midrule
% DFT & 54.91 & 13.27\% \\
% \rowcolor[gray]{0.95} PDSA-WB (ours) & 54.16 & 15.22\% \\
% \rowcolor[gray]{0.95} PDSA (ours) & 55.69 & 15.45\% \\
% \bottomrule
% \end{tabular}
% \end{table}

% 需要 booktabs + colortbl
\begin{table}[!ht]
\vspace{-5ex}
\small
\setlength{\aboverulesep}{0pt}
\setlength{\belowrulesep}{0pt}
\setlength{\extrarowheight}{0pt}
\centering
\setlength{\tabcolsep}{3pt}
\renewcommand{\arraystretch}{0.95}
\caption{Driving safety evaluation.}
\vspace{-2ex}
\label{tab:driving_ability_side}

\begin{tabular}{l|cc|cc}
\toprule

 & \multicolumn{2}{c|}{\textbf{Person2Drive}} 
 & \multicolumn{2}{c}{\textbf{Bench2Drive}} \\
\midrule
\textbf{Method} 
& Driving Score$\uparrow$ 
& Success Rate$\uparrow$ 
& Driving Score$\uparrow$ 
& Success Rate$\uparrow$ \\
\midrule
Baseline 
& 58.22 
& 98.9\% 
& 54.91 
& 13.27\% \\

\rowcolor[gray]{0.95}
PDSA-WB 
& 67.45
& 98.6\%
& 54.16 
& 15.22\% \\

\rowcolor[gray]{0.95}
\textbf{PDSA}
& \textbf{70.13}
& 98.4\%
& \textbf{55.69} 
& \textbf{15.45\%} \\

\bottomrule
\end{tabular}
\vspace{-6ex}
\end{table}

% 表6的表头是生成的，要改！！！

% More analysis and discussion can be found in the supplement.
% 这句是之前写的，还能怎么补充？！

\vspace{-1ex}
\subsection{Fine-tuning Trajectory Visualization}
% To better verify whether our proposed PDSA can align the driving styles, we conduct trajectory visualization here. Specifically, we provide the visualization of ground truth, the trajectory before fine-tuning, and the trajectory after fine-tuning with driver individuals in our dataset. The results are shown in Figure~\ref{fig:plt1} and Figure~\ref{fig:plt2}.

\begin{figure}[ht]
\vspace{-3ex}
    \centering
\includegraphics[width=1\linewidth]{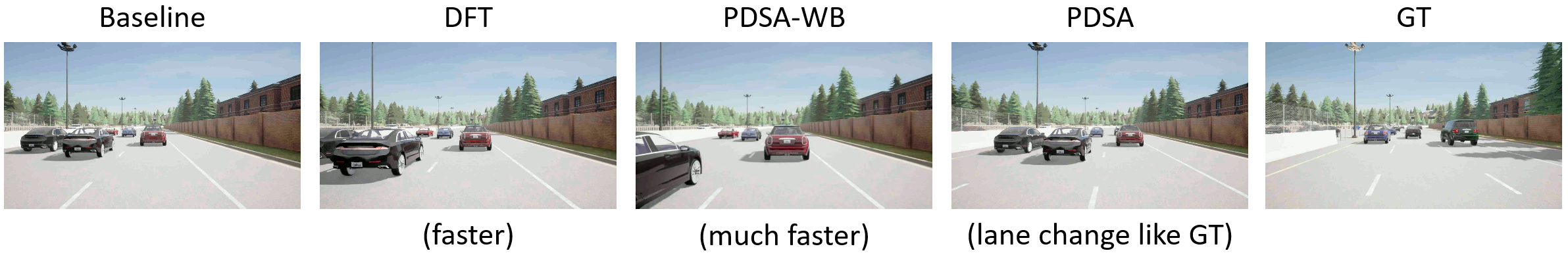}
    \vspace{-6mm}
    \caption{
        Comparison of driving behaviors across different models, visualized from the vehicle’s first-person view.
        %While other variants mainly learn to drive faster, only PDSA successfully reproduces the human driver’s lane-change behavior, yielding trajectories that most closely match the ground-truth driving style.
    }
    \label{fig:plt1}
    \vspace{-5ex}
\end{figure}

\begin{figure}[ht]
\vspace{-5ex}
    \centering
\includegraphics[width=1\linewidth]{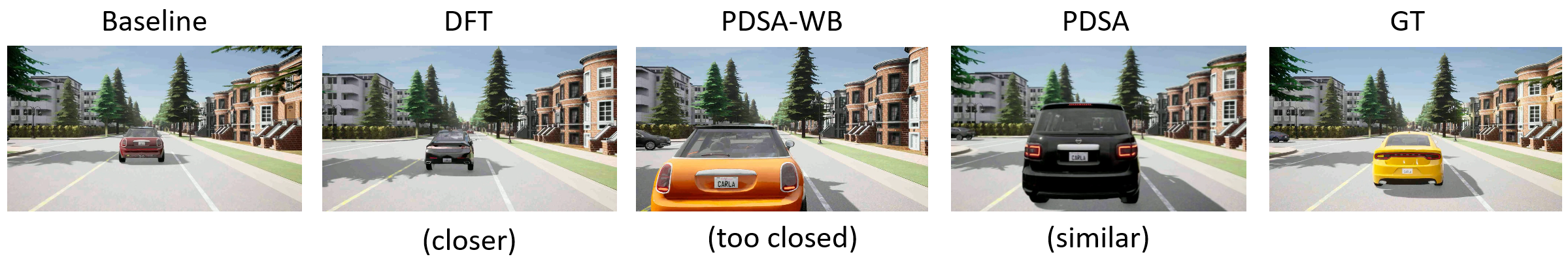}
    \vspace{-6mm}
    \caption{
        Comparison of post-stop following gaps across different models, visualized from the vehicle’s first-person view.
        %Baseline and DFT exhibit noticeable deviations from human ground truth, either stopping too close or too far from the lead vehicle. In contrast, PDSA achieves the closest alignment with human post-stop following behavior, demonstrating the strongest ability to reproduce individual stopping style.
    }
    \label{fig:plt2}
    \vspace{-4ex}
\end{figure}

To further validate the effectiveness of our PDSA method, we randomly sample a subset of instances and provide qualitative visualizations across scenarios. The selected lane-changing and post-stop following examples are shown in Figure~\ref{fig:plt1} and Figure~\ref{fig:plt2}. 
Figure~\ref{fig:plt1} shows that, in contrast to the other model variants—which primarily exhibit a tendency to drive faster—PDSA is the only method that successfully reproduces the human driver’s lane-change behavior. It captures both the timing and execution of the maneuver, producing trajectories that align most closely with the ground-truth driving style. 
Figure~\ref{fig:plt2} shows that baseline and DFT show clear discrepancies from human ground truth, often stopping either too close to or too far from the leading vehicle. In comparison, PDSA produces stopping behaviors that align most closely with human drivers, demonstrating its superior capability to capture and reproduce individualized post-stop following style. 
Overall, the visualization confirms the quantitative results: PDSA effectively adapts the model toward each driver’s unique style, achieving noticeable convergence in both motion dynamics and trajectory geometry.

% \begin{figure}[t]
%     \centering
% \includegraphics[width=0.5\linewidth]{Fig/visualization/vis4.jpg}
%     \vspace{-5mm}
%     \caption{
%         Comparison between baseline,gt and fine-tuned, showing that fine-tuning helps capture personalized driving styles.
%     }
%     \label{fig:driving_vis}
%     \vspace{-3ex}
% \end{figure}

\vspace{-3ex}
\section{Conclusion}
\vspace{-1ex}
In this work, we address the gap in achieving truly individual-level personalized  E2E-AD. To this end, we develop an open-source and extensible data collection platform and construct Person2Drive. This large-scale personalized driving dataset enables fine-grained style analysis across multiple drivers and scenarios. Building upon this foundation, we introduce an explainable evaluation system for quantitative driving style assessment, establishing the first unified standard for evaluating individual driving styles. Furthermore, we propose a style-guided lightweight fine-tuning framework that allows autonomous models to generate personalized yet robust trajectories. Comprehensive experiments and analyses on both Person2Drive and public datasets validate the effectiveness of our framework. We believe this work provides a unified benchmark and methodological foundation for future research in user-aware and trustworthy autonomous driving, paving the way toward truly human-centered E2E-AD.
\vspace{-1ex}

\section*{Acknowledgements }
\noindent This work was supported by the National Natural Science Foundation of China (Project Number 62595774), MoE Key Laboratory of Intelligent Perception and Human-Machine Collaboration (KLIP-HuMaCo), HPC Platform of ShanghaiTech University.

% \clearpage
\bibliographystyle{splncs04}
\bibliography{main}
\end{document}